\pdfoutput=1
\documentclass[11pt]{article}

\usepackage[]{acl}

\usepackage{times}
\usepackage{latexsym}
\usepackage{float}
\usepackage{graphicx}
\usepackage{tabularx}
\usepackage{xcolor,colortbl}
\usepackage{dirtytalk}

\usepackage[T1]{fontenc}
\usepackage[utf8]{inputenc}

\usepackage{microtype}

\title{Differential Privacy in Natural Language Processing: The Story So Far}%, Current Limitations, and Open Directions}

\author{Oleksandra Klymenko, Stephen Meisenbacher \and Florian Matthes \\
        Technical University of Munich \\ Department of Informatics \\ %Software Engineering for Business Information Systems \\ 
       Garching, Germany \\
        \texttt{\{alexandra.klymenko, stephen.meisenbacher, matthes\}@tum.de}}

\begin{document}

\maketitle

\begin{abstract}
As the tide of Big Data continues to influence the landscape of Natural Language Processing (NLP), the utilization of modern NLP methods has grounded itself in this data, in order to tackle a variety of text-based tasks. These methods without a doubt can include private or otherwise personally identifiable information. As such, the question of privacy in NLP has gained fervor in recent years, coinciding with the development of new Privacy-Enhancing Technologies (PETs). Among these PETs, Differential Privacy boasts several desirable qualities in the conversation surrounding data privacy. Naturally, the question becomes whether Differential Privacy is applicable in the largely unstructured realm of NLP. This topic has sparked novel research, which is unified in one basic goal: how can one adapt Differential Privacy to NLP methods? This paper aims to summarize the vulnerabilities addressed by Differential Privacy, the current thinking, and above all, the crucial next steps that must be considered.
\end{abstract}

\section{Introduction}
\label{intro}
In an age where a vast amount of data is being produced daily, the opportunities created by this proliferation increase concurrently. The availability of big data enables countless downstream tasks whose accuracy and utility seem to increase with the amount of data used. Specifically, the fields of Machine Learning (ML) and Deep Learning (DL) have profited from such data. Particularly in the case of Natural Language Processing (NLP), the tasks at hand more often than not concern the handling of \textit{unstructured} data, meaning data that is not neatly organized into a traditional row-like database structure, and furthermore, data that is not necessarily static. In fact, it is estimated that data on the order of zettabytes (ZB) is being produced every day \cite{8399125}, and within this amount, roughly 80\% is unstructured, e.g. textual data \cite{hammoud2019personal}. %Indeed, this results in quite a baffling amount of unstructured data.

At the same time as this profound boom in popularity of big data tasks, there has been an increase in the attention paid to the way in which data is used, specifically to the issue of \textit{privacy}. The problem is exacerbated when sensitive parts of the data relate to a specific task (e.g. with medical data). The threat becomes more serious when the models themselves used with the learning tasks are vulnerable to attacks.

Although many useful Privacy-Enhancing Technologies have emerged, one in particular seems to be a good fit when faced with the scale of these big data learning tasks: Differential Privacy \cite{dwork2006differential}. The key feature of Differential Privacy is its mathematically grounded notion of privacy, which can be intuitively explained using the privacy parameter, most often called $\epsilon$. This idea was originally intended for data stored in structured databases, i.e. a relational schema. As a result, Differential Privacy upon its inception became an excellent way to start to reason about privacy in ML and DL models that were trained on these types of databases.

Alas, in the field of NLP, where the core unit of data is unstructured, fuzzy text rather than a structured data point, an initial attempt to apply Differential Privacy poses some challenges. Chief among these is the challenge of how to transfer the core concepts of Differential Privacy, namely the \say{individual} and \textit{adjacency}, to the textual domain where these concepts are not easily perceivable. Thus, it becomes the goal to find new ways of reasoning about Differential Privacy in order to adapt it to the unstructured data domain of NLP. Through the course of this paper, the foundations of Differential Privacy in the lens of NLP will be investigated, motivated by some privacy vulnerabilities that surface from NLP techniques. Afterwards, the limitations and open questions of Differential Privacy with NLP will be analyzed with an in-depth discussion.

\section{Foundations}
\paragraph*{Privacy-Enhancing Technologies}
Several PETs have been created with the goal of protecting the privacy of the individuals. Three methods in particular have arisen as useful ways to reason about groups in a dataset: \textit{k-anonymity} \cite{Samarati1998ProtectingPW}, \textit{l-diversity} \cite{10.1145/1217299.1217302}, and \textit{t-closeness} \cite{4221659}. These frameworks are quite reliant upon the structured nature of a database, yet they become impractical in the realm of large-scale, unstructured data. They therefore lack a reasonable applicability to NLP. Addressing privacy concerns within text, traditional methods include simple redaction or scrubbing based upon available heuristics. Newer notions, such as \textit{t-plausibility} \cite{5283278}, were designed with text document sanitization in mind. Finally, modern approaches involve the idea of \textit{adversarial learning}, such as \cite{elazar-goldberg-2018-adversarial} or \cite{friedrich-etal-2019-adversarial}. As one may postulate, Differential Privacy also lacks a direct mapping to NLP, becoming the basis of investigation in the pursuit of differentially private NLP.

\paragraph*{Differential Privacy in ML and DL}
Researchers first looked to determine the place of Differential Privacy in ML and DL. The following papers on Differential Privacy in ML \cite{ji2014differential} and DL \cite{Abadi_2016} are great starting points for applying Differential Privacy to these areas. Importantly, it has been shown that Differential Privacy does indeed have a place when considering these types of learning tasks. Not until later was the idea extended to NLP, and even today, the research on it is still relatively scarce. This is due precisely to some of the reasons introduced in Section \ref{intro}. Nevertheless, this extra layer of complexity makes Differential Privacy in NLP an interesting topic. There exist papers that systematize this topic for ML, such as \cite{8677282}, and DL \cite{ppdp}, which partially cover Differential Privacy, but to the best of the authors' knowledge, no such papers specifically address its application to NLP. Thus, it becomes the goal to start to bridge this gap.

\section{Methodology}
To accomplish the goals of this paper, the following research questions have been defined:

\begin{enumerate}
    \itemsep0em
    \item[RQ1] What vulnerabilities to NLP techniques is Differential Privacy capable of preventing?
    \item[RQ2] What is the current state of Differential Privacy in its application to NLP?
    \item[RQ3] What are the predominant current limitations and future directions of applying Differential Privacy to NLP?
\end{enumerate}

The structure of the research supporting this paper is twofold, firstly taking the form of a systematic literature review. Thus, the main method of answering the stated research questions will be to seek out relevant academic literature and research, which will serve as the primary source for data synthesis.  This process, including formulating a search process and creating exclusion criteria, is based upon Garousi \cite{GAROUSI2019101}. %Figure \ref{method} illustrates the complete methodology followed.

%\begin{figure}[ht]
%    \centering
%    \includegraphics[width=0.45\textwidth]{methodology5.png}
%    \vspace{-5pt}
%    \caption{Proposed Methodology}
%    \label{method}
%\end{figure}

%Ultimately, the systematic literature review process will provide readers with a clearly written and understandable motivation for the application of Differential Privacy to NLP (in defense against privacy risks), and furthermore, how and to what degree privacy preservation can be reasoned about.

The second stage of research involves conducting semi-structured expert interviews. The main goal of these is to supplement the knowledge gained from the literature with practical viewpoints from privacy professionals and relevant academic researchers. This is crucial to harmonizing the promise of research with the demands of industry, and ultimately, society. Table \ref{tab:int_table} shows a summary of the four interviews conducted. The insights from these interviews will be highlighted in the discussion conducted in Section \ref{discuss}.

\begin{table}[ht]
  \centering
    %\resizebox{1\columnwidth}{!}{
    \begin{tabular}{|p{0.1\linewidth}p{0.35\linewidth}p{0.35\linewidth}|}
    \hline
    \textbf{Code} & \textbf{Position} & \textbf{Organization} \\
    \hline
    I1    & Co-Founder and CEO & Privacy-focused AI startup, Canada \\
    \hline
    I2    & Postdoctoral Research Associate & University, Australia\\
    \hline
    I3    & Applied Science Manager & Research division of large American tech company \\
    \hline
    I4    & PhD Candidate & University, USA \\
    \hline
    \end{tabular}%
    %}
  \caption{Coded Interviewee Table}
  \label{tab:int_table}%
  \vspace{-10pt}
\end{table}%

The remainder of this paper is structured as follows: Section \ref{priv} begins our exploration of Differential Privacy in NLP by first analyzing which privacy vulnerabilities Differential Privacy is best suited to address (RQ1). Next, Section \ref{dp1} introduces Differential Privacy in the scope of how it has been adapted to textual data (RQ2). Section \ref{dp2} continues this narrative by focusing on a generalization of Differential Privacy that is well-suited for unstructured domains (RQ2). Finally, Section \ref{discuss} comprises of several discussion points that are seen to be pertinent current limitations, and accordingly, crucial future research directions (RQ3).

\section{Privacy Vulnerabilities in NLP Techniques}
\label{priv}
By first analyzing some privacy vulnerabilities in NLP techniques (RQ1), we hope to motivate the thinking behind the incorporation of Differential Privacy in NLP, presented in Sections \ref{dp1} and \ref{dp2}. Here, we differentiate between two overarching categories of vulnerabilties: (1) \textit{information leakage} \cite{song2020information} and (2) \textit{unintended memorization} \cite{carlini2019secret}. The focus is placed on the former, as this is more relevant to NLP, while the latter pertains more generally to the DL applications.

\subsection{Language Leakage}
When approaching any number of NLP tasks, the first step ultimately becomes finding an appropriate text representation. An early, simple example of this would be the Bag-of-Words model, or representing text by a set of linguistic-based features. This kind of modeling, however, also enables the building of a \say{stylometric profile}. In the wrong hands, the collection of these features can give up \textit{implicit information}, which is not explicitly sensitive but can highlight user (author) attributes. This type of hidden information is known as \textit{information leakage}, but in light of the focus on textual data, we use the term \textit{language leakage}. Such a generalization aids in seeing that both traditional and more modern (i.e. embedding) representations of text are susceptible to such leakage.

In recent years, the growing success of word embeddings for use as general purpose language models has rooted their utilization in downstream tasks. The usefulness of these models lies in the fact that numerical representations of textual data can be used for computation in a wide variety of learning tasks, where plaintext does not readily fit. Also inherent to these models are useful properties that can capture word associations. In order to create them, word embeddings are usually trained on vast amounts of text. These texts could contain private or sensitive information, which in turn are encoded into the vector representations. This poses a problem with embeddings, whose goal is to capture semantic meaning of words, without an inherent concept of private information.

Beyond embeddings, the rising ubiquity of (large) language models, or (L)LMs, such as GPT-2/3, has called to question Differential Privacy's role in this domain. For similar reasons as embeddings, LMs trained on massive amounts of textual data are susceptible to leakage of sensitive information contained therein. As such, it becomes the task to incorporate Differential Privacy into these LMs to defend against inference attacks, while still preserving their utility.

\subsubsection{Exploitation}
When thinking about the components of text that may comprise sensitive information, one may imagine that much of this follows a structured, fixed format. Examples of this include, but are not limited to, Social Security numbers (SSNs), birth dates, and phone numbers. When textual data contains such structure, it can become the goal of an attacker to recover, or reconstruct, these fixed-formatted strings. Such attacks have been shown to be effective by \cite{Pan2020PrivacyRO} and \cite{carlini2020extracting}, especially when certain embedding models are utilized. As such, exploiting language leakage within text representations generally revolves around \textit{inference}.

Keyword inference attacks present a more general attack model, where the attacker has an idea of what kind of text is contained in the released data. Concretely, the attacker's goal is to extract keywords from the data, given some domain knowledge. It is shown in \cite{Pan2020PrivacyRO} that keyword extraction is also possible where the attacker has little to no domain knowledge of the data.

In addition to extracting information about input data in embedding models, the authors in \cite{song2020information} demonstrate the ability to extract author attributes. Furthermore, the structure of embedding models is susceptible to leaking membership information, especially with infrequently occurring inputs to the embedding model. Similar results concerning the inference of author attributes come out of \cite{coavoux2018privacypreserving}.

Alarmingly, it has been shown that even a simple combination of lexical and syntactic features can be used to predict the gender of a text’s author with approximately 80\% accuracy \cite{gender} - and this is done with a relatively simple, non-neural classifier. Other similar cases are covered in \cite{elazar-goldberg-2018-adversarial}. One might imagine how such features can not only expose author attributes, but also the author's identity.

\subsection{Unintended Memorization}
\label{mem}
As the prevalence of neural NLP has been on the rise in recent years, concerns about the ability of neural networks to memorize data, or rather the patterns therein, has lead to questions of privacy breaches. In \cite{carlini2019secret}, it is shown that a relatively rare-occurring \textit{secret} in the training text can cause a neural model to memorize it completely. In some cases, such memorization seems to be a necessary part of the training process. The authors in \cite{thomas2020investigating} show that certain word embedding models, when used in neural networks, lead to unintended memorization. Although solutions to this problem involve Differential Privacy \cite{Abadi_2016, carlini2019secret, DBLP:journals/corr/abs-2102-12677}, it is not the main focus of this paper.

\subsection{Risk Use Cases}
We discuss two general categories of risk use cases Differential Privacy in NLP can address, as well as imply when it is not appropriate.

\subsubsection{Data Release}
Often, it might make sense to release (unstructured) textual data to third parties. In many of these cases, however, the text being released contains sensitive information. A well-studied example of this is the release of medical data, which can take the form of hospital records or doctors' notes \cite{Li2018ProtectingPW}. Other prevalent use cases include the release of text from online reviews, social media posts, or government records \cite{Pan2020PrivacyRO}, all of which can contain quite sensitive information. For data release, such data is often transferred to third parties in de-identified form \cite{Abdalla2020UsingWE}, with the thought that this inherently provides a first layer of defense. Even so, a malicious user with access can extract personal information, showcased in \cite{info:doi/10.2196/18055}, which shows that releasing medical data in embedding form still allows for nearly 70\% reconstruction of Personally Identifiable Information. 

\subsubsection{Model Abuse}\label{modelabuse}
Many modern NLP techniques utilize some neural component, often in combination with embedding representations. In some of these cases, users interact with the models dynamically. Two broad categories of this interaction are: (1) centralized learning, in which users upload data to a centralized model for computations, and (2) decentralized (collaborative) learning, where computation is done locally with updates from a central server. If a malicious user has a point of access to either of these types of systems, information about the data can be inferred based on two ways \cite{9044259}: (1) black-box access, where the malicious user can query the model an unlimited number of times, and thus gain information from the model outputs, and (2) white-box access, where there is access to the original model parameters.

%\subsection{Summary of Privacy Vulnerabilities} \label{exploit}
%With these privacy vulnerabilities existing in NLP, and with some use cases in mind, one can now see from the previous sections how these vulnerabilities might be exploited so as to learn some private information. These particular attacks revolve around \textit{inference} attacks, using publicly released parts of the model to infer information about the data used to train them. The focus of this paper is placed here, since Differential Privacy is most readily equipped to address such vulnerabilities. 

\begin{figure}[htbp]
    \centering
    \vspace{2pt}
    \includegraphics[width=0.45\textwidth]{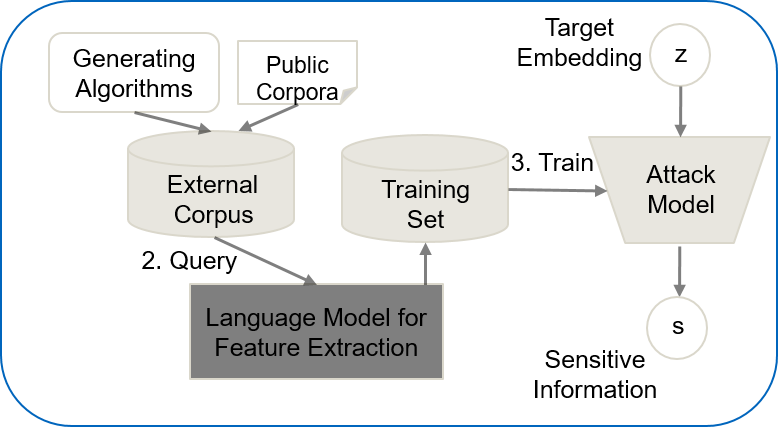}
    \caption{General Attack Pipeline, based on \cite{Pan2020PrivacyRO}}
    \label{attack}
    \vspace{-8pt}
\end{figure}

\subsection{General Attack Pipeline}
In order to define a general attack pipeline on NLP models as defined in both \cite{Pan2020PrivacyRO} and \cite{lyu2020differentially}, a few assumptions must be made about the attacker: (1) the attacker has access to the target text representations or model, (2) the attacker knows which pre-trained language model was used, and (3) the attacker is able to recreate the text representation. Note that these assumptions can be generalized to any target text representation, including plaintext. The assumptions enable the formulation of a general attack pipeline, illustrated in Figure \ref{attack}. It enables an attacker in possession of sensitive text data encoded into some representation to \textit{infer} the contents within. 
This idea of inference becomes the crucial basis to where Differential Privacy comes into play.

\section{Differential Privacy in NLP}
\label{dp1}
With the privacy issues that can arise when performing NLP tasks in mind, it is a logical step to consider the application of Differential Privacy to mitigate these privacy issues. Before one can consider \textit{how} to do this, it may be be useful to understand \textit{what} exactly Differential Privacy can protect against in the context of NLP.

\subsection{Differential Privacy}
\label{DP}
Differential Privacy \cite{dwork2006differential} was first proposed with the goal of approaching privacy-preservation by protecting the \textit{individual} in a database, and doing so with a mathematical guarantee. The underlying idea of \textit{randomized response} is transferred to Differential Privacy by saying that the result of some query on two exactly identical databases except for one individual is similar within some threshold, defined by the privacy parameter $\epsilon$, or the \textit{privacy budget}. The exact foundations are covered briefly next, but one may refer to \cite{Wood2018DifferentialPA} for a thorough primer.

\subsection{Foundations}
The idea of Differential Privacy revolves around the protection of the individual in a database, or dataset. Traditionally, the \say{individual} being referred to corresponds to a single data entry, representing one individual's information structured according to the database's schema. With this in mind, the definition of Differential Privacy is expressed as the following inequality:
\begin{equation} \label{eq1}
    Pr[\mathcal{K}(D) \in S] \le e^\epsilon Pr[\mathcal{K}(D') \in S]
\end{equation}
The first important aspect to note in Equation \ref{eq1} is that the output of some model is probabilistic, governed by some randomized function $\mathcal{K}$. Within this system, $\mathcal{K}$ has a possible set of outputs given an input database, denoted by $S$, where $S \subseteq \textrm{Range}(\mathcal{K})$. To make this concrete, given a database $D$ as input to $\mathcal{K}$, $S$ comprises of the values that can be returned as an output. Next, Eq. \ref{eq1} refers to two \textit{neighboring} databases $D$ and $D'$, which according to Differential Privacy, are two databases which differ in exactly one element, or more precisely one individual (\textit{Hamming Distance} of 1). In effect, this means that any two databases which are identical minus one element are indeed \textit{adjacent}, i.e. fit the description of $D$ and $D'$. As a final component, Eq. \ref{eq1} includes $e^\epsilon$ as a bound of how much the output of two adjacent datasets can differ, with $\epsilon$ as the \textit{privacy parameter}. Intuitively, one can see that with a lower $\epsilon$, the two outputs are constrained to be more similar, and on the flip side, a larger $\epsilon$ provides a bit more leeway. With this definition, the concept of \textit{indistinguishably} is given form, with $\epsilon$ controlling how indistinguishable, or not, these operations on two neighboring databases must be. With a chosen $\epsilon$, it is said that a function $\mathcal{K}$ achieves $\epsilon$-differential privacy if Equation \ref{eq1} is satisfied.

As one can see, this definition provides a quantifiable way to envision privacy in datasets, bolstered by a flexible privacy parameter. Translating this notion to the unstructured textual domain, though, comes with its challenges. Before these are discussed, one must first analyze how exactly Differential Privacy may be beneficial for privacy preservation in NLP, and in what way.

\subsection{Protection \textit{Against} Inferences?}
Section \ref{priv} introduced some ways in which attackers can possibly gain sensitive information from text-based data, which revolve around the ability to \textit{infer} information. When considering Differential Privacy as a potential defense for these attacks, it is important to notice that it does not protect against inferences themselves -- and this applies to the application of Differential Privacy to any domain. In other words, a differentially private system is still vulnerable to inference attacks.

What Differential Privacy does offer, however, refers back to its core concept: protection of the \textit{individual} against inferences. With NLP, that is with unstructured text data, this must be reasoned about differently. The application of Differential Privacy to the NLP domain would mean to provide the individuals (data contributors) plausible deniability as a protection against inference attacks. Put more concretely, one can take the example of keyword inference. Although an attacker still might be able to infer keywords from text representations, there would exist a level of uncertainty as to whether this extracted keyword actually represents the true, original keyword. As a result, the privacy protection given by Differential Privacy is rooted in this sense of plausible deniability, and not by a complete protection against inferences themselves.

\subsection{The Challenge with Unstructured Data}
\label{uns}
Of course, the notion of the \say{individual} in a structured dataset is not immediately transferable to a non-structured dataset, such as a corpus of text (documents), yet this can be accomplished somewhat easily by reasoning about the individuals whose data is contained within such a corpus. With this thought, however, the concept of a \say{database} becomes unclear -- is a database a collection of documents each tied to an individual, or is a database a single document comprised of many \textit{individual words}? In the former case, applying Differential Privacy becomes difficult without a way to define adjacency beyond the traditional Hamming Distance. Likewise, the latter case would result in a very strict (and not practical) constraint.

The solution to applying standard Differential Privacy (i.e. in its original form) to NLP comes by converting text to a latent representation, and subsequently applying some differentially private mechanism. The biggest challenge, and seeming shortcoming, of such an approach is that using Differential Privacy in its original form imposes quite strict constraints in terms of how to perturb a given piece of text. Ultimately, this means that one must consider any two text documents to be adjacent, much like in the way that any two entries in a structured dataset are neighboring, thus taking a very conservative view of adjacency for text. A direct answer to this challenge comes with a generalized notion, introduced in Section \ref{dp2}.

\subsection{Applications}
\label{dpapp}
Several implementations have appeared in the literature, all of which leverage Differential Privacy in the context of NLP tasks. As such, the following works represent the current thinking of how Differential Privacy can be used in practice for NLP.

In \cite{lyu2020differentially}, a method is proposed to perturb binary vector text representations in a simple, yet differentially private manner. \cite{weggenmann2018syntf} focuses on TF-IDF vectors, leveraging the Exponential Mechanism \cite{mcsherry2007mechanism} to create \say{synthetic} vectors. The authors in \cite{bo2019erae} add on an embedding reward system to encourage a diversity in the output text. \cite{beigi2019i} also approach the utility vs. privacy problem with the introduction of a discriminator in a two-autoencoder setup.

In light of several works applying Differentially Private Stochastic Gradient Descent (DP-SGD) to address the memorization issue in deep neural NLP models (see Section \ref{mem}), the authors in \cite{DBLP:journals/corr/abs-2110-06500} instead address privacy in the underlying language models. Here, differentially private fine-tuning is performed on several popular LMs.

Others focus on leveraging Differential Privacy in specific tasks, such as n-gram extraction \cite{https://doi.org/10.48550/arxiv.2108.02831}, topic modeling \cite{9671552}, or financial text classification \cite{https://doi.org/10.48550/arxiv.2110.01643}.

An interesting case comes with \cite{DBLP:journals/corr/abs-2102-01502}, whose implementation is later refuted by the author of \cite{habernal-2021-differential}. Similarly, Habernal \cite{https://doi.org/10.48550/arxiv.2202.12138} claims that the DPText implementation of \cite{beigi2019i} fails to be differentially private. This becomes the basis of an important discussion in Section \ref{explain}.

\section{Metric Differential Privacy for NLP}
\label{dp2}
The idea of $d_\mathcal{X}$-privacy \cite{Chatzikokolakis2013BroadeningTS}, also $d$-privacy or Metric Differential Privacy, was first introduced in 2013 as a generalization of Differential Privacy, with the goal of extending the concept beyond structured databases to arbitrary domains (e.g. location data). The key for achieving this comes with the reasoning about \textit{adjacency} between two databases. In domains without an immediate notion of adjacency between individuals, it becomes necessary to find an alternate expression. The answer comes with the utilization of a (distance) metric existing within some \textit{metric space}, whose members are often referred to as \textit{points}. A relaxed sense of Differential Privacy thus enables its application to arbitrary domains endowed with a metric, and naturally this fits well with text.

\subsection{Foundations}
With an available metric, one can say that the distinguishability between two databases imposed by Differential Privacy depends on the distance between, or similarity of, these two databases. Therefore, the smaller the distance (and greater the similarity), the more similar (indistinguishable) the output of some function on the two databases must be. One can see that this is an extension of \say{differing by one individual} to \say{differing by some value}. With this in mind, the original Equation \ref{eq1} is adapted to fit this thinking, yielding:
\begin{equation} \label{eq4}
     Pr[\mathcal{K}(x) \in S] \le e^{\epsilon d(x,x')} Pr[\mathcal{K}(x') \in S]
\end{equation}
The implications of the new Equation \ref{eq4} become clear: as the metric value between two inputs becomes larger (i.e. the inputs are less related), the distinguishability between the outputs resulting from them is allowed to be greater, and vice versa.

The task is now to apply the concepts of $d_\mathcal{X}$-privacy directly to NLP techniques utilizing text representations. It is important to note that there exist several other generalizations of Differential Privacy, as systematized in \cite{DBLP:journals/corr/abs-1906-01337}, yet the focus here is placed on $d_\mathcal{X}$-privacy due to its direct applicability to NLP tasks.

The main difference brought by the introduction of $d_\mathcal{X}$-privacy to NLP comes with the direct incorporation of a metric that \say{scales} the noise addition process to achieve Differential Privacy. In short: more similar meaning $\rightarrow$ more required indistinguishability. This new aspect comes as very convenient when dealing with text representations that already exist within spaces endowed with a distance (similarity) metric. $d_\mathcal{X}$-privacy allows for an increased flexibility in the sense that the underlying basis for a text representation (e.g. Euclidean vs. Hyperbolic) can change, without affecting the Differential Privacy inequality or compromising privacy preservation. This will prove to be useful as novel text representation methods are introduced.

\subsection{Applications}
\label{dapp}
Early approaches \cite{fernandes2018author, fernandes2019generalised, feyisetan2019privacy} involved working within the Euclidean space, i.e. using $n$-dimensional embeddings and the Laplace Mechanism. In \cite{feyisetan2019leveraging}, a shift to hyperbolic space was performed to model the hierarchical relationships within a language, leveraging them to perturb text. Finally, \cite{xu2020differentially} makes the switch to the Mahalanobis (elliptical) norm which takes into account the shape of a particular space, resulting in better perturbation of sparse words. In a recent implementation \cite{DBLP:journals/corr/abs-2107-07928}, a bridge between Differential Privacy and Metric Differential Privacy is created through the use of a \say{Truncated Exponential Mechanism}.

These works encapsulate the current thinking as to how $d_\mathcal{X}$-privacy can be implemented with the NLP models of today. One might imagine, however, that $d_\mathcal{X}$-privacy is not presently widely utilized due to its relative adolescence. 

\section{Discussion}
\label{discuss}
With the application of Differential Privacy to the area of NLP also come several challenges. Ultimately, these limitations serve as a basis for future work and motivation for further improvements.

\subsection{Utility}
\label{util}
One would certainly be remiss to discuss the topic of Privacy-Enhancing Technologies without addressing the ever-present privacy-utility tradeoff. With this topic come many interesting findings from the literature, which are not necessarily all negative. With this said, the flip side of the coin presents an arguably more pressing discussion point. The usual effect is that as the $\epsilon$ parameter is set to be lower (stricter), the accuracy of a given task clearly decreases. Although this may be discouraging news, one must keep in mind that there is \say{no free lunch}. The implications of this in terms of applying Differential Privacy to NLP, then, varies from case to case: one needs to decide to what degree privacy is necessary. I1 illustrates this complex decision in real-world applications by saying, \say{it’s hard because yes your accuracy is lower if you use Differential Privacy, but if you don’t use it you wouldn’t get access to the data in the first place}. The bright side comes from the flexibility that Differential Privacy offers. Adjusting $\epsilon$ enables one to experiment with the privacy and utility results of various parameters.

\subsection{Benchmarking}
Along with this current limitation of utility surfaces a clear lack in the present literature: benchmarking. The original works themselves and even dedicated papers such as \cite{DBLP:journals/corr/abs-2106-13973} often present findings regarding utility in the form of established scoring schemes (accuracy, F1). However, other important aspects of utility, especially in the mindset of NLP, are often ignored. Above all, the ability for these Differential Privacy implementations to produce coherent, grammatically correct language is often left out. One such paper, \cite{bo2019erae}, does make this attempt, yet the results are not too convincing utility-wise. Therefore, a greater focus on syntactical and semantic coherence, sentence flow, and readability is needed.

Another aspect of benchmarking that is completely absent in the literature is the computational power, i.e. resources and time, required to implement the proposed methods. In order to make Differential Privacy for NLP a viable option going forward, more work on this will be required. Moreover, the question of transparency goes hand-in-hand with that of explainability, discussed in Section \ref{explain}.

\subsection{Structural Limitations}
The key to reasoning about Differential Privacy in the unstructured domain of language comes with the important step of imposing a sort of \say{quasi-structure}, e.g. by reasoning about text representations. This raises the question: is such a transfer of concepts always necessary when applying Differential Privacy? It was shown what happens when one attempts to deviate a bit from the rigorous definition put forth by Differential Privacy, specifically in the form of $d_\mathcal{X}$-privacy applying to arbitrary domains. Using $d_\mathcal{X}$-privacy as a case study, it becomes interesting to see how much one can diverge from the original sense of Differential Privacy to fit the needs of increasingly unstructured domains. 

This becomes even more pertinent when addressing one of the major assumptions made throughout the literature, which is that the databases in question, whether structured or not, are \textit{static} in nature. The notion that a database is static and does not evolve over time is indeed fitting with the original purpose and definition of Differential Privacy, yet it is less and less representative of a major part of the data being produced today \cite{stream}. As a result, there now exists a discrepancy between the basis for proposed applications of Differential Privacy to NLP and what is used in state-of-the-art NLP. I3 states the problem more concretely: 
\begin{quote}
    You have this beautiful theory, these nice robust proofs, all of the protection against side attacks and post-processing, compositionality, all of these lovely things... then you say something like you have an epsilon budget of 2 and it will be refreshed every 4 days, then the whole thing becomes meaningless at that point!
\end{quote}
An investigation into this matter was started in \cite{DBLP:journals/corr/abs-1803-06416}, and one more tailored to NLP surely needs to be conducted going forward.

Both with standard Differential Privacy and $d_\mathcal{X}$-privacy, the general approach so far in the literature is to (1) calculate some latent representation, (2) apply noise, and (3) proceed \say{downstream}. The observed effect as shown in the literature has its flaws: the output after the noise addition often results in less than optimal language, with an overall lack of natural flow (also covered in Section \ref{util}). 

Another current bottleneck that arises from these implications is the reliance on word embedding models. I3 calls this \say{the big elephant in the room}. In earlier models where the corresponding embeddings are calculated based upon co-occurrence, the application of Differential Privacy makes more sense: perturbation results in semantically related noisy outputs. Recently, though, the utilization of contextual word embeddings (e.g. BERT) has become the prevalent method, and this presents a problem for the current thinking with Differential Privacy in NLP. With contextual embeddings, noise addition followed by a projection will result not in semantically similar words, but rather contextually similar ones -- this is not desired for meaning- and utility-preserving private text representations. In essence, \say{with contextual embeddings, you would no longer be able to compute your nearest neighbor index, and [current Differential Privacy] becomes an impossibility} (I3).

\subsection{Context}
Beyond the problem posed by Differential Privacy with contextual text representations, the idea of \textit{context} raises further questions. In the realm of textual data, the notion of what may be considered \say{private} presumably is quite dependent on the context in which this text was created or expressed, such as with customer reviews versus medical records. Even beyond this, the fact that \textit{privacy} is an incredibly personal (and cultural) notion makes seemingly rigid definitions, such as that of Differential Privacy, hard to reason about. In this light, perhaps the idea of societal context must be investigated and incorporated in regards to text, so that differentially private NLP becomes more relevant. A related discussion built upon this idea follows in the ensuing section.

\subsection{Explainability}
\label{explain}
Possibly one of the more crucial points that one must consider when applying Differential Privacy to NLP is the notion of explainability. The main question is: at what point is text truly private? 

This question presents the biggest challenge to better explainability. At the core of the challenge lies the issue of \textit{what exactly} it is about text that needs to become private. Of course, there could exist explicit words or phrases that contain sensitive information. Going deeper, though, one can also consider \textit{stylometry} as a threat: our \textit{writing style} is inherently personal. As pointed out by I1: 
\begin{quote}
The one thing with NLP that you won’t get with a machine learning community is a deeper understanding of the language – what might be sensitive in the language, so things like an understanding of all the things you can learn from language – who is writing something, their profile – so having a more cohesive understanding of what is happening with text.
\end{quote}
I2 also adds: \say{First thing we need to ask: is there really a privacy issue? What is the privacy issue? Can you demonstrate it?} With these questions in mind, the interesting aspect that comes with differentially private NLP is that the input text itself, or rather the text representations, are being perturbed, in contrast to operating on structured databases. This begs another question: how does perturbing word $x$ and mapping it to word $y$ increase the privacy protection of some individual? Another important design decision that seems to be ignored so far involves the so-called \textit{selection problem}. In the literature, this issue is usually handled via the way in which text can be perturbed, or mapped, to other semantically similar text. The flip side of this coin, \textit{selecting} what parts or sections of text are private and need to be handled accordingly, has received little to no recent attention. All of these questions are introduced when there is rarely a structured or direct mapping of database entries to individuals.

For a clearer answer, one can look to the crucial $\epsilon$ parameter. I4 supports this in saying, \say{We don’t have a formal definition of privacy, but I think this mathematical guarantee has made it easier for us to work with privacy}. This, however, turns out to be at the heart of the explainability issue. On one side, it allows for a relative quantification of privacy with respect to the value of the parameter. The challenging part is that this $\epsilon$ does not immediately lend itself to a clear path for explaining privacy in NLP. Even if this were possible, the literature seems to vary in terms of what $\epsilon$ makes sense for a given application of Differential Privacy to NLP, suggesting that the $\epsilon$ parameter might indeed just be relative to the task at hand. And as I2 formulates it, \say{the down side to [Differential Privacy] is that there is not a really strong operational interpretation of what privacy means}. In this case, $\epsilon$ loses its global explainability value a bit, or rather, its \say{operational interpretation}.

A final matter falling under the umbrella of explainability is the relative shroud of mystery surrounding Differential Privacy. Even amongst researchers, there seems to be a confusion of how to apply it correctly, as demonstrated by \cite{habernal-2021-differential} and \cite{https://doi.org/10.48550/arxiv.2202.12138}. As Habernal points out, the crux of the issue lies in the fact that \say{it seems non-trivial to get [Differential Privacy] right when applying it to NLP}. The promise of Differential Privacy may be quite enticing, but as I1 puts it, \say{you have to get someone who understands the technology properly and understands the privacy-preserving nature}. One can extrapolate from here and assume that explaining the mechanisms (and merits) of Differential Privacy to the general public will be a complex task. Accordingly, more emphasis on education and awareness should be afforded.

\subsection{Future Directions: A Summary}
The possibilities for future work relating to the application of Differential Privacy to NLP have been alluded to throughout and discussed via limitations in Section \ref{discuss}, but they are made explicit here:
\begin{itemize}
    \itemsep0em
    \item The \textbf{continued exploration of the privacy-utility tradeoff} when using Differential Privacy in NLP, as well as better explaining it.
    \item The \textbf{integration of Differential Privacy in more modern NLP architectures}, particularly sequence models, e.g. transformers.
    \item A focus on making Differential Privacy \textbf{compatible and usable with more recent text representations} (e.g. contextual embeddings and LLMs).
    \item The investigation of \textbf{Differential Privacy's role, applicability, and effectiveness in non-static data settings}: in particular, reasoning about how it could work with streaming (text) datasets.
    \item The topic of $d_\mathcal{X}$-privacy opens the doors to \textbf{other possible generalizations of Differential Privacy} tailored to NLP.
    \item Differential Privacy, NLP, and their \textbf{relation to regulation, policy, and implementation} in practice.
    \item \textbf{The ability to explain Differential Privacy} and its role in NLP, conducting research \say{in a way that people can understand} (I4).
\end{itemize}

\section{Conclusion}
The investigation into Differential Privacy's place within the NLP sphere results in many interesting findings and discussions. Understanding that there does indeed exist privacy vulnerabilities to NLP techniques, looking to Differential Privacy for a solution does not come without its challenges. Above all, this requires additional consideration as to how some core privacy concepts translate to the underlying structure (or lack thereof) powering current NLP tasks. The theoretical foundations and applications arising from recent literature have provided an excellent initial excursion into this topic, and from them, one can derive promising avenues for future improvements. Where Differential Privacy in NLP goes from here is yet to be seen, but the primary goal of this paper was to explore its foundations and to start the discussion on what this future might look like. Ultimately, the promise of applying Differential Privacy to mitigate privacy issues in NLP places it on the vanguard of Privacy-Enhancing Technologies, demanding further research.

\section*{Acknowledgements}
This work has been supported by the German Federal Ministry of Education and Research (BMBF)
Software Campus grant LACE 01IS17049 and the Bavarian Research Institute for Digital Transformation (bidt).

\bibliography{acl_latex}
\bibliographystyle{acl_natbib}

%\appendix

%\section{Example Appendix}
%\label{sec:appendix}

%This is an appendix.

\end{document}